\documentclass[conference]{IEEEtran}
\usepackage{cite}

\ifCLASSINFOpdf
  \usepackage[pdftex]{graphicx}
  \DeclareGraphicsExtensions{.pdf,.jpeg,.png}
\else
   \usepackage[dvips]{graphicx}
   \DeclareGraphicsExtensions{.eps}
\fi
\usepackage{amsmath}
\usepackage[linesnumbered,ruled]{algorithm2e}
\usepackage{array}
\usepackage[caption=false,font=footnotesize]{subfig}
\usepackage{stfloats}
\usepackage{microtype}
\usepackage{url}
\usepackage{xcolor,colortbl}
\definecolor{Gray}{gray}{0.85}
\usepackage{enumerate}

\begin{document}

\title{Evolving Deep Convolutional Neural Networks \\ for Image Classification}

\author{\IEEEauthorblockN{Yanan Sun,
Bing Xue, Mengjie Zhang and Gary G. Yen}
}

\maketitle

\begin{abstract}
Evolutionary computation methods have been successfully applied to neural networks since two decades ago, while those methods cannot scale well to the modern deep neural networks due to the complicated architectures and large quantities of connection weights. In this paper, we propose a new method using genetic algorithms for evolving the architectures and connection weight initialization values of a deep convolutional neural network to address image classification problems. In the proposed algorithm, an efficient variable-length gene encoding strategy is designed to represent the different building blocks and the unpredictable optimal depth in convolutional neural networks. In addition, a new representation scheme is developed for effectively initializing connection weights of deep convolutional neural networks, which is expected to avoid networks getting stuck into local minima which is typically a major issue in the backward gradient-based optimization. Furthermore, a novel fitness evaluation method is proposed to speed up the heuristic search with substantially less computational resource. The proposed algorithm is examined and compared with 22 existing algorithms on nine widely used image classification tasks, including the state-of-the-art methods. The experimental results demonstrate the remarkable superiority of the proposed algorithm over the state-of-the-art algorithms in terms of classification error rate and the number of parameters (weights).
\end{abstract}

\begin{IEEEkeywords}
Genetic algorithms, convolutional neural network, image classification, deep learning.
\end{IEEEkeywords}

\IEEEpeerreviewmaketitle

\section{Introduction}
\label{section_1}
\IEEEPARstart{C}{onvolutional} neural networks (CNNs) have demonstrated their exceptional superiority in visual recognition tasks, such as traffic sign recognition~\cite{cirecsan2012multi}, biological image segmentation~\cite{ning2005toward}, and image classification~\cite{krizhevsky2012imagenet}. CNNs are originally motivated by the computational model of the cat visual cortex specializing in processing vision and signal related tasks~\cite{hubel1962receptive}. Since LetNet-5 was proposed in 1989~\cite{lecun1989backpropagation,lecun1998gradient}, which is an implementation of CNNs and whose connection weights are optimized by the Back-Propagation (BP) algorithm~\cite{rumelhart1988learning}, various variants of CNNs have been developed, such as VGGNet~\cite{simonyan2014very} and ResNet~\cite{he2015deep}. These variants significantly improve the classification accuracies of the best rivals in image classification tasks. Diverse variants of CNNs differ from their architectures and weight connections.

Mathematically, a CNN can be formulated by~(\ref{equ_cnn}) in the context of an image classification task with the input $(X,Y)$,
\begin{equation}
\label{equ_cnn}
\left\{
\begin{array}{l}
A_{rchitecture} = F(X, Y) \\
W_{eight} = G(A_{rchitecture})\\
minimize~~L(X, W_{eight}, Y)
\end{array}
\right.
\end{equation}
where $X$ and $Y$ are the input data and corresponding label, respectively, $F(\cdot)$ denotes the architecture choosing function with the given data, $G(\cdot)$ refers to the initialization method of the connection weights $W_{eight}$ based on the chosen architecture, and $L(\cdot)$ measures the differences between the true label and the label predicted by the CNN using $X$ and $W_{eight}$. Typically, the Gradient Descend (GD)-based approaches, e.g., Stochastic GD (SGD), are utilized to minimize $L(X, W_{eight}, Y)$ within the given number of epochs, where the connection weight values are iteratively updated. Although $L(\cdot)$ is not differentiable in all occasions, GD-based methods are preferred due to their effectiveness and good scalability as the number of connection weights increases. A CNN commonly has a huge number of connection weights. However, $F(\cdot)$ and $G(\cdot)$ are countable functions that are discrete and neither convex or concave, and they are not well addressed in practice. Furthermore, because the gradient-based optimizers are heavily dependent on the initial values of the weights (including biases), it is essential to choose a suitable $G(\cdot)$ that can help the consecutive GD-based approaches to escape from local minima. Furthermore, the performance of assigned architectures cannot be evaluated until the minimization of $L(\cdot)$ is finished, while the minimization is a progress of multiple iterations, which in turn increases the difficulty of choosing the potential optimal $F(\cdot)$. Therefore, the architecture design and connection weight initialization strategy should be carefully treated in CNNs.

Typically, most of the architecture design approaches were initially developed for the deep learning algorithms in the early date (e.g., the Stacked Auto-Encoders (SAEs)~\cite{bourlard1988auto,hinton1994autoencoders} and the Deep Belief Networks (DBNs)~\cite{hinton2006reducing}), such as Grid Search (GS), Random Search (RS)~\cite{bergstra2012random}, Bayesian-based Gaussion Process (BGP)~\cite{rasmussen2006gaussian,movckus1975bayesian}, Tree-structured Parzen Estimators (TPE)~\cite{bergstra2011algorithms}, and Sequential Model-Based Global Optimization (SMBO)~\cite{hutter2011sequential}. Theoretically, GS exhaustively tests all combinations of the parameters to expectedly seize the best one. However, GS cannot evaluate all combinations within an acceptable time in reality. Moreover, GS is difficult to optimize the parameters of continuous values~\cite{bergstra2012random}. RS could reduce the exhaustive adverse of GS, but the absolute ``random'' severely challenges the sampling behavior in the search space~\cite{beck1981integer,halton1964radical,atanassov2009tuning}. In addition, BGP incurs extra parameters (i.e., the kernels) that are arduous to tune. TPE treats each parameter independently, while the most key parameters in CNNs are dependent (e.g., the convolutional layer size and its stride, more details can be seen in Subsection~\ref{section_2_intro_cnn}). The methods mentioned above have shown their good performance in most SAEs and DBNs, but are not suitable to CNNs. Their success in SAEs and DBNs is due largely to the architecture construction approaches, which are greedy layer-wised by stacking a group of building blocks with the same structures (i.e., the three-layer neural networks). In each building block, these architecture-search methods are utilized for only optimizing the parameters, such as the number of neurons in the corresponding layer. However in CNNs, the layer-wised method cannot be applied due to their architecture characteristics of non-stacking routine, and we must confirm the entire architectures at a time. Furthermore, multiple different building blocks exist in CNNs, and different orders of them would result in significantly different performance. Therefore, the architecture design in CNNs should be carefully treated. Recently, Baker \textit{et al.}~\cite{baker2017designing} proposed an architecture design approach for CNNs based on reinforcement learning, named MetaQNN, which employed 10 Graphic Processing Unit (GPU) cards with 8-10 days for the experiments on the CIFAR-10 dataset.

Due to the drawbacks of existing methods and limited computational resources available to interested researchers, most of these works in CNNs are typically performed by experts with rich domain knowledge~\cite{bergstra2012random}. Genetic Algorithms (GAs), which are a paradigm of the evolutionary algorithms that do not require rich domain knowledge~\cite{ashlock2006evolutionary,back1996evolutionary}, adapt the meta-heuristic pattern motivated by the process of natural selection~\cite{mitchell1998introduction} for optimization problems. GAs are preferred in various fields due to their characteristics of gradient-free and insensitivity to local minima~\cite{yao1999evolving}. These promising properties are collectively achieved by a repeated series of the selection, mutation, and crossover operators. Therefore, it can be naturally utilized for the optimization of architecture design and the connection weight initialization for CNNs. Indeed, GAs for evolving neural networks can be traced back to 1989~\cite{miller1989designing}. In 1999, Yao~\cite{yao1999evolving} presented a survey about these different approaches, which are largely for the optimization of connection weights in the fixed architecture topologies. In 2002, Stanley and Miikkulainen proposed the Neuron-Evolution Augmenting Topology (NEAT)~\cite{stanley2002evolving} algorithm to evolve the architecture and connection weights of a small scale neural network. Afterwards, the HyperNEAT~\cite{stanley2009hypercube}, i.e., NEAT combined with the compositional pattern producing networks~\cite{stanley2007compositional}, was proposed to evolve a larger scale neural network with an indirect encoding strategy. Motivated by the HyperNEAT, multiple variants~\cite{pugh2013evolving,kim2015deep,fernando2016convolution} were proposed to evolve even larger scale neural networks. However, the major deficiencies of the HyperNEAT-based approaches are: 1) they are only suitable for evolving deep neural networks with global connection and single building blocks, such as SAEs and DBNs, but not CNNs where local connection exists and multiple different building blocks need to be evolved simultaneously, and 2) hybrid weigh connections (e.g., connections between two layers that are non-adjacent) may be evolved, which are contrast to the architectures of CNNs. Indeed, the views have been many years that evolutionary algorithms are incapable of evolving the architecture and connection weights in CNNs due to the tremendous number of related parameters~\cite{zoph2016neural,baker2016designing,verbancsics2013generative}. Until very recently in 2017, Google showed their Large Evolution for Image Classification (LEIC) method specializing at the architecture optimization of CNNs~\cite{real2017large}. LETC is materialized by GAs without the crossover operator, implemented on 250 high-end computers, and archives competitive performance against the state-of-the-art on the CIFAR-10 dataset by training for about 20 days. Actually, by directly using GAs for the architecture design of CNNs, several issues would raise in nature: 1) the best architecture is unknown until the performance is received based on it. However, evaluating the performance of one individual takes a long time, and appears more severely for the entire population. This would require much more computational resources for speeding up the evolution; 2) the optimal depth of CNNs for one particular problem is unknown, therefore it is hard to constrain the search space for the architecture optimization. In this regard, a variable-length gene encoding strategy may be the best choice for both 1) and 2), but how to assign the crossover operation for different building blocks is a newly resulted problem; and 3) the weight initialization values heavily affect the performance of the confirmed architecture, but addressing this problem involves a good gene encoding strategy and the optimization of hundreds and thousands decision variables.

\subsection{Goal}
\label{sec_goal}
The aim of this paper is to design and develop an effective and efficient GA method to automatically discover good architectures and corresponding connection weight initialization values of CNNs (i.e., the first two formulae in~(\ref{equ_cnn})) without manual intervention. To achieve this goal, the objectives below have been specified:
\begin{enumerate}
	\item Design a flexible gene encoding scheme of the architecture, which does not constrain the maximal length of the building blocks in CNNs. With this gene encoding scheme, the evolved architecture is expected to benefit CNNs to achieve good performance in solving different tasks at hand.
	\item Investigate the connection weight encoding strategy, which is capable of representing tremendous numbers of the connection weights in an economy way. With this encoding approach, the weight connection initialization problem in CNNs is expected to be effectively optimized by the proposed GA.
	\item Develop associated selection (including the environmental selection), crossover, and mutation operators that can cope with the designed gene encoding strategies of both architectures and connection weights.
	\item Propose an effective fitness measure of the individuals representing different CNNs, which does not require intensive computational resources.
	\item Investigate whether the new approach significantly outperform the existing methods in both classification accuracy and number of weights.
\end{enumerate}

\subsection{Organization}
The reminder of this paper is organized as follows: the background of the CNNs, the related works on the architecture design and weight initialization approaches of CNNs are reviewed in Section~\ref{section_2}. The framework and the details of each step in the proposed algorithm are elaborated in Section~\ref{section_3}. The experiment design and experimental results of the proposed algorithm are shown in Sections~\ref{section_4} and~\ref{sec_experimental_results}, respectively. Next, further discussions are made in Section~\ref{section_5}. Finally, the conclusions and future work are detailed in Section~\ref{section_6}.

\section{Background and Related Work}
\label{section_2}

\subsection{Architecture of Convolutional Neural Network}
\label{section_2_intro_cnn}
\begin{figure}
	\centering
	\includegraphics[width=\columnwidth]{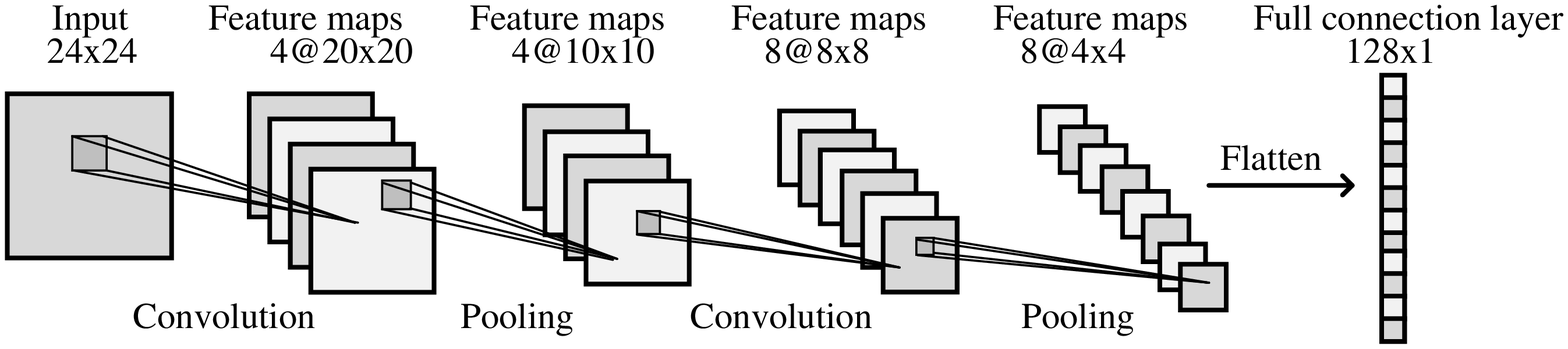}
	\caption{An general architecture of the Convolutional Neural Network.}
	\label{fig_cnn}
\end{figure}
\subsubsection{\textbf{Bone of CNNs}} Fig.~\ref{fig_cnn} exhibits an extensive architecture of one CNN, where there are two convolutional operations, two pooling operations, the resulted four groups of feature maps, and the full connection layer in the tail. The last layer, which is a full connection layer, receives the input data by flattening all elements of the fourth group of feature maps. Generally, the convolutional layers and the pooling layers can be mixed to stack together in the head of the architecture, while the full connection layers are constantly stacked with each other in the tail of the architecture. The numbers in Fig.~\ref{fig_cnn} refer to the sizes of the corresponding layer. Particularly, the input data is with $24\times 24$, the output is with $128\times 1$, and the other numbers denote the feature map configurations. For example, $4@20\times 20$ implies there are $4$ feature maps, each with the size of $4\times 4$.

In the following, the details of the convolutional layer and the pooling layer, which are associated with the convolution and the pooling operations, respectively, are documented in detail, while the full connection layer is not intended to describe here because it is well-known.

\subsubsection{\textbf{Convolution}} Given an input image with the size of $n\times n$, in order to receive a feature map generated by the convolutional operations, a filter must be defined in advance. Actually, a filter (it can also be simply seen as a matrix) is randomly initialized with a predefined size (i.e., the filter width and the filter height). Then, this filter travels from the leftmost to the rightmost of the input data with the step size equal to a stride (i.e., the stride width), and then travels again after moving downward with the step size equal to a stride (i.e., the stride height), until reach the right bottom of the input image. Depending on whether to keep the same sizes between the feature map and the input data through padding zeros, the convolutional operations are categorized into two types: the VALID (without padding) and the SAME (with padding). Specifically, each element in the feature map is the sum of the products of each element from the filter and the corresponding elements this filter overlaps. If the input data is with multiple channels, say $3$, one feature map will also require $3$ different filters, and each filter convolves on each channel, then the results are summed element-wised.

\begin{figure}[htp]
	\centering
	\includegraphics[width=\columnwidth]{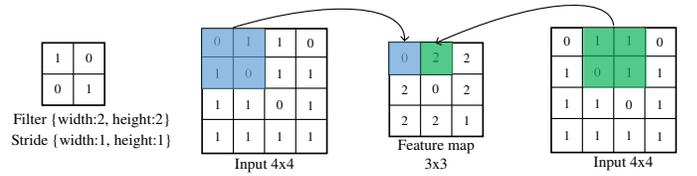}
	\caption{An illustration of convolutional operation.}
	\label{fig_convolution}
\end{figure}

An example of the VALID convolutional operation is illustrated by Fig.~\ref{fig_convolution}, where the filter has the same width and height equal to $2$, the input data is with the size of $4\times 4$, and the stride has the same height and width equal to $1$. As shown in Fig.~\ref{fig_convolution}, the generated feature map is with the size of $3\times 3$, the shadow areas in the input data with different colors refer to the overlaps with the filter at different positions of the input data, the shadow areas in the feature map are the respective resulted convolutional outcomes, and numbers in the filter are the connection weight values. Generally, convolutional results regarding each filter are updated by adding a bias term and then through a nonlinearity, such as the Rectifier Linear Unit (ReLU)~\cite{glorot2011deep}, before they are stored into the feature map. Obviously, the involved parameters in one convolutional operation are the \textit{filter width}, the \textit{filter height}, the \textit{number of feature maps}, the \textit{stride width}, the \textit{stride height}, the \textit{convolutional type}, and the \textit{connection weight} in the filter.

\subsubsection{\textbf{Pooling}} 
\begin{figure}[htp]
	\centering
	\includegraphics[width=\columnwidth]{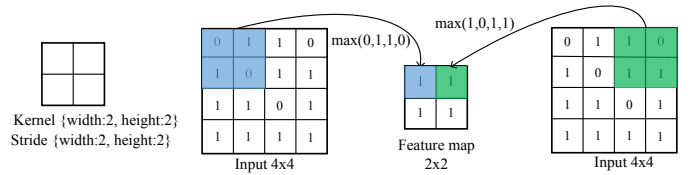}
	\caption{An illustration of pooling operation.}
	\label{fig_pooling}
\end{figure}

Intuitively, the pooling operation resembles the convolution operation except for the element-wised product and the resulted values of the corresponding feature map. Briefly, the pooling operation employs a predefined window (i.e., the kernel) to collect the average value or the maximal value of the elements where it slides, and the slide size is also called ``stride'' as in the convolutional operation. For better understanding, an example of the pooling operation is illustrated by Fig.~\ref{fig_pooling}, where the kernel is with the size of $2\times 2$, and the both stride width and height are $2$, the input data is with the size of $4\times 4$, the shadows with different colors refer to the two slide positions and the resulted pooling values. In this example, the maximal pooling operation is employed. Evidently in the pooling operation, the involved parameters are the \textit{kernel width}, the \textit{kernel height}, the \textit{stride width}, the \textit{stride height}, and the \textit{pooling type}.

\subsection{CNN Architecture Design Algorithm}
In this subsection, only LEIC~\cite{real2017large} is concerned due to its specific intention for the architecture design of CNNs. To this end, we will review the algorithm and then point out the limitations, which are used to highlight the necessity of the corresponding work in the proposed algorithm.

LEIC employed a GA to evolve the architectures of CNNs, where individuals were evolved from scratch, and each individual was encoded with a variable-length chromosome. During the phase of evolution, different kinds of layers can be incorporated into the individuals by the mutation operation, with the expectation that individuals with promising performance will be generated. Note that, the crossover operation, which are designed for local search in GAs, were not investigated in the main part of LEIC. Without any crossover operator, GAs typically require a large population size for reducing the adverse impact. Therefore, LETC adopted a population size of $10^3$, while that of a magnitude with order $10^2$ is a general setting in the GA community. In addition, $250$ high-end computers were employed for LEIC, which was caused by the fitness assignment approach utilized in LEIC. In LEIC, each individual was evaluated with the final image classification accuracy, which typically took a long time due to the iterative nature of GD-based optimizers. However, LEIC reported an external experiment on the crossover operation, which was used only for exchanging the mutation probabilities and the connection weights trained by SGD. Although it is difficult to reach the actual reason why the crossover has not been used for evolving the architectures in LETC, it is obvious at least that the crossover operations are not easy to achieve for chromosomes with different lengths. This would be more complicated in CNNs that have multiple different building blocks.

In summary, the main deficiency of LEIC is its high computational complexity mainly caused by the fitness evaluation strategy it adopts as well as without crossover operations. The huge computational resource that LEIC requires makes it very intractable in academic environment.

\subsection{Connection Weight Initialization}
\label{subsection_weight_init_method}
Typically, the initialization methods are classified into three different categorises. The first employs constant numbers to initialize all connection weights, such as the zero initializer, one initializer, and other fixed value initializer. The second is the distribution initializer, such as using the Gaussion distribution or uniform distribution to initialize the weights. The third covers the initialization approaches with some prior knowledge, and the famous Xavier initializer~\cite{glorot2010understanding} belongs to this category. Because of the numerous connection weights existing in CNNs, it is not necessary that all the connection weights start with the same values, which is the major deficiency of the first initialization method. In the second initialization approach, the shortage of the first one has been solved, but the major difficulty exists in choosing the parameter of the distribution, such as the range of the uniform distribution, and the mean value as well as the standard derivation of the Gaussian distribution. To solve this problem, the Xavier initializer presented a range for uniform sampling based on the neuron saturation prior using the sigmoid activation function~\cite{hodgkin1952quantitative}. Supposed the number of neurons in two adjacent layers are $n_1$ and $n_2$, the values of the weights connecting these two layers are initialized within the range of $\left[-\sqrt{6/(n_1+n_2)}, \sqrt{6/(n_1+n_2)}\right]$ by uniformly sampling. Although the Xavier initializer works better than the initialization methods from the other two categories, a couple of major issues exist: 1) It highly relies on the architectures of CNNs, particularity the number of neurons in each layer in the networks (e.g., $n_1$ and $n_2$ in its formulation). If the optimal architectures of the networks are not found, the resulted initialized parameters perform badly as well, then there is no way to evaluate the desired performance of the architectures and may mislead the adjustment of the architectures. 2) The Xavier initializer is presented on the usage of the sigmoid activation, while the widely used activation function in CNNs is the ReLU~\cite{krizhevsky2012imagenet,zeiler2014visualizing,szegedy2015going,he2015deep}.

To the best of our knowledge, there has not been any existing evolutionary algorithm for searching for the connection weight initialization of deep learning algorithms including CNNs. The main reason is the tremendous numbers of weights, which are difficult to be effectively encoded into the chromosomes and to be efficiently optimized due to its large-scale global optimization nature~\cite{omidvar2014cooperative}.

\section{The proposed algorithm}
\label{section_3}
In this section, the proposed Evolving deep Convolutional Neural Networks (EvoCNN) for image classification is documented in detail.
\subsection{Algorithm Overview}
\label{sec_algorithm_overview}
\begin{algorithm}
	\label{alg_framework}
	\caption{Framework of EvoCNN}
	$P_0 \leftarrow$Initialize the population with the proposed gene encoding strategy;\\
	\label{alg_line_1}
	$t\leftarrow 0$;\\
	\label{alg_line_2}
	\While{termination criterion is not satisfied}
	{\label{alg_begin_evo}
		Evaluate the fitness of individuals in $P_t$;\\
		\label{alg_line_4}
		$S\leftarrow$ Select parent solutions with the developed slack binary tournament selection;\\
		\label{alg_line_4_5}
		$Q_t\leftarrow$ Generate offsprings with the designed genetic operators from $S$;\\
		\label{alg_line_5}
		$P_{t+1}\leftarrow$Environmental selection from $P_t\cup Q_t$;\\
		\label{alg_line_6}
		$t\leftarrow t+1$;
	}\label{alg_end_evo}
	Select the best individual from $P_t$ and decode it to the corresponding convolutional neural network.
	\label{alg_line_9}
\end{algorithm}

Algorithm~\ref{alg_framework} outlines the framework of the proposed EvoCNN method. Firstly, the population is initialized based on the proposed flexible gene encoding strategy (line~\ref{alg_line_1}). Then, the evolution begins to take effect until a predefined termination criterion, such as the maximum number of the generations, has been satisfied (lines~\ref{alg_begin_evo}-\ref{alg_end_evo}). Finally, the best individual is selected and decoded to the corresponding CNN (line~\ref{alg_line_9}) for final training.

During the evolution, all individuals are evaluated first based on the proposed efficient fitness measurement (line~\ref{alg_line_4}). After that, parent solutions are selected by the developed slack binary tournament selection (line~\ref{alg_line_4_5}), and new offspring are generated with the designed genetic operators (line~\ref{alg_line_5}). Next, representatives are selected from the existing individuals and the generated offsprings to form the population in the next generation to participate subsequent evolution (line~\ref{alg_line_6}). In the following subsections, the key steps in Algorithm~\ref{alg_framework} are narrated in detail.

\subsection{Gene Encoding Strategy}
\label{section_3_1}

\begin{figure}
	\centering
	\includegraphics[width=0.8\columnwidth]{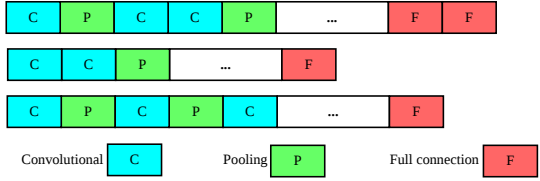}
	\caption{An example of three chromosomes with different lengths in EvoCNN.}
	\label{fig_chromosome}
\end{figure}

\begin{table}
	\renewcommand{\arraystretch}{1.1}
	\caption{The encoded information in EvoCNN.}
	\label{tab_encoded_info}
	\begin{center}
		\begin{tabular}{p{0.2\columnwidth}<{\centering}|p{0.7\columnwidth}}
			\hline
			Unit Type & \multicolumn{1}{c}{Encoded Information} \\
			\hline
			convolutional layer & the filter width, the filter height, the number of feature maps, the stride width, the stride height, the convolutional type, the standard deviation and the mean value of filter elements\\
			\hline
			pooling layer& the kernel width, the kernel height, the stride width, the stride height, and the pooling type (i.e., the average or the maximal)\\
			\hline
			full connection layer & the number of neurons, the standard deviation of connection weights, and the mean value of connection weights\\
			\hline
		\end{tabular}
	\end{center}
\end{table}

As introduced in Subsection~\ref{section_2_intro_cnn}, three different building blocks, i.e., the \textit{convolutional} layer, the \textit{pooling} layer, and the \textit{full connection} layer, exist in the architectures of CNNs. Therefore, they should be encoded in parallel into one chromosome for evolution. Because the optimal depth of a CNN in solving one particular problem is unknown prior to confirming its architecture, the variable-length gene encoding strategy, which is very suitable for this occasion, is employed in the proposed EvoCNN method. Furthermore, because the performance of CNNs is highly affected by their depths~\cite{bengio2013representation,bengio2011expressive,delalleau2011shallow,simonyan2014very}, this variable-length gene encoding strategy makes the proposed EvoCNN method have chances to reach the best result due to no constrains on the architecture search space.

In particular, an example of three chromosomes with different lengths from EvoCNN is illustrated by Fig.~\ref{fig_chromosome}, and all represented information in these three layers are detailed in Table~\ref{tab_encoded_info}. Commonly, hundreds of thousands connection weights may exist in one convolutional or full connection layer, which cannot be all explicitly represented by a chromosome and effectively optimized by GAs. Therefore, in EvoCNN, we use only two statistical real numbers, the standard derivation and mean value of the connection weights, to represent the numerous weight parameters, which can be easily implemented by GAs. When the optimal mean value and the standard derivation are received, the connection weights are sampled from the corresponding Gaussian distribution. The details of population initialization in EvoCNN are given in the next subsection based on the gene encoding strategy introduced above.

\subsection{Population Initialization}
\label{section_3_2}
\begin{algorithm}
	\label{alg_population_init}
	\caption{Population Initialization}
	\KwIn{The population size $N$, the maximal number of convolutional and pooling layers $N_{cp}$, and the maximal number of full connection layers $N_{f}$.}
	\KwOut{Initialized population $P_0$.}
	$P_0\leftarrow\emptyset$;\\
	\While{$|P_0| \leq N$}
	{
		$part_1\leftarrow \emptyset$;\\
		\label{alg_init_pop_part1_begin}
		$n_{cp}\leftarrow$ Uniformaly generate an integer between $[1, N_{cp}]$;\\
		\While{$|part_1| \leq n_{cp}$}
		{
			$r\leftarrow$ Uniformly generated a number between $[0, 1]$;\\
			\uIf{$r\leq 0.5$}
			{\label{alg_init_pop_tossing}
				$l\leftarrow$ Initialize a convolutional layer with random settings;\\
			}
			\Else{
				$l\leftarrow$ Initialize a pooling layer with random settings;\\
			}
			$part_1\leftarrow part_1\cup l$;
		}\label{alg_init_pop_part1_end}

		$part_2\leftarrow \emptyset$;\\
		\label{alg_init_pop_part2_begin}
		$n_{f}\leftarrow$ Uniformaly generate an integer between $[1, N_{f}]$;\\
		\While{$|part_2| \leq n_{f}$}
		{
			$l\leftarrow$ Initialize a full connection layer with random settings;\\
			$part_2\leftarrow part_2\cup l$;
		}\label{alg_init_pop_part2_end}
		$P_0\leftarrow P_0\cup (part_1\cup part_2)$;\\
	}
	\textbf{Return} $P_0$.
\end{algorithm}
 For convenience of the elaboration, each chromosome is separated into two parts. The first part includes the convolutional layers and the pooling layers, and the other part is the full connection layers. Based on the convention of the CNN architectures, the first part starts with one convolutional layer. The second part can be added to only at the tail of the first part. In addition, the length of each part is set by randomly choosing a number within a predefined range.

Algorithm~\ref{alg_population_init} shows the major steps of the population initialization, where $|\cdot|$ is a cardinality operator, lines~\ref{alg_init_pop_part1_begin}-\ref{alg_init_pop_part1_end} show the generation of the first part of one chromosome, and lines~\ref{alg_init_pop_part2_begin}-\ref{alg_init_pop_part2_end} show that of the second part. During the initialization of the first part, a convolutional layer is added first. Then, a convolutional layer or a pooling layer is determined by the once coin tossing probability and then added to the end, which is repeated until the predefined length of this part is met. For the second part, full connection layers are chosen and then added. Note here that, convolutional layers, pooling layers, and full connection layers are initialized with the random settings, i.e., the information encoded into them are randomly specified before they are stored into the corresponding part. After these two parts finished, they are combined and returned as one chromosome. With the same approach, a population of individuals are generated.

\subsection{Fitness Evaluation}
\label{section_3_3}
\begin{algorithm}
	\label{alg_fitness_evaluation}
	\caption{Fitness Evaluation}
	\KwIn{The population $P_t$, the training epoch number $k$ for measuring the accuracy tendency, the training set $D_{train}$, the fitness evaluation dataset $D_{fitness}$, and the batch size $num\_of\_batch$.}
	\KwOut{The population with fitness $P_t$.}
	\For{each individual $s$ in $P_t$}
	{
		$i\leftarrow 1$;\\
		$eval\_steps\leftarrow |D_{fitness}|/num\_of\_batch$;\\
		\While{$i\leq k$}
		{
			Train the connection weights of the CNN represented by individual $s$;\\
			\If{$i == k$}
			{\label{alg_fitness_evaluation_acc_begin}
				$accy\_list\leftarrow \emptyset$;\\
				$j\leftarrow 1$;\\
				\While{$j\leq eval\_steps$}
				{
					$accy_j\leftarrow$ Evaluate the classification error on the $j$-th batch data from $D_{fitness}$;\\
					$accy\_list\leftarrow accy\_list\cup accy_j$;\\
					$j\leftarrow j + 1$;
				}
				Calculate the number of parameters in $s$, the mean value and standard derivation from $accy\_list$, assign them to individual $s$, and update $s$ from $P_t$;\\
			}
			\label{alg_fitness_evaluation_acc_end}

			$i\leftarrow i+1$;
		}
	}
	\textbf{Return} $P_t$.
\end{algorithm}

Fitness evaluation aims at giving a quantitative measure determining which individuals qualify for serving as parent solutions. Algorithm~\ref{alg_fitness_evaluation} manifests the framework of the fitness evaluation in EvoCNN.
Because EvoCNN concerns on solving image classification tasks, the classification error is the best strategy to assign their fitness. The number of connection weights is also chosen as an additional indicator to measure the individual's quality based on the principle of Occam's razor~\cite{blumer1987occam}.

 With the conventions, each represented CNN is trained on the training set $D_{train}$, and the fitness is estimated on another dataset $D_{fitness}$\footnote{The original training set is randomly split into $D_{train}$ and $D_{fitness}$, where $D_{fitness}$ is unseen to the CNN training phase, which can give a good indication of the generalization accuracy on the test set.}. CNNs are frequently with deep architectures, thus thoroughly training them for receiving the final classification error would take considerable expenditure of computing resource and a very long time due to the large number of training epochs required ($>$100 epochs are invariably treated in fully training CNNs). This will make it much more impracticable here due to the population-based GAs with multiple generations (i.e., each individual will take a full training in each generation). We have designed an efficient method to address this concern. In this method, each individual is trained with only a small number of epochs, say $5$ or $10$ epochs, based on their architectures and connection weight initialization values, and then the mean value and the standard derivation of classification error are calculated on each batch of $D_{fitness}$ in the last epoch. Both the mean value and the standard derivation of classification errors, are employed as the fitness of one individual. Obviously, the smaller mean value, the better individual. When the compared individuals are with the same mean values, the less standard derivation indicates the better one.

In summary, three indicators are used in the fitness evaluation, which are the mean value, standard derivation, and the number of parameters. There are several motivations behind this fitness evaluation strategy: 1) It is sufficient to investigating only the tendency of the performance. If individuals are with better performance in the first several training epochs of CNNs, they will probably still have the better performance in the following training epochs with greater confidence. 2) The mean value and the standard derivation are statistical significance indicators, thus suitable for investigating this tendency, and the final classification error can be received by optimizing only the best individual evolved by the proposed EvoCNN method. 3) CNN models with less number of connection weights are preferred by smart devices (more details are discussed in Section~\ref{section_5}).

\subsection{Slack Binary Tournament Selection}
\label{section_3_binary_selection}
\begin{algorithm}
	\label{alg_tournament_selection}
	\caption{Slack Binary Tournament Selection}
	\KwIn{Two compared individuals, the mean value threshold $\alpha$, and the paramemter number threshold $\beta$.}
	\KwOut{The selected individual.}
	$s_1\leftarrow$ The individual with larger mean value;\\
	$s_2\leftarrow$ The other individual;\\
	$\mu_1, \mu_2\leftarrow$ The mean values of $s_1, s_2$;\\
	$std_1, std_2\leftarrow$ The standard derivations of $s_1, s_2$;\\
	$c_1, c_2\leftarrow$ The parameter numbers of $s_1, s_2$;\\
	\uIf{$\mu_1-\mu_2>\alpha$}
	{
		\textbf{Return} $s_1$.
	}\Else{
		\uIf{$c_1-c_2>\beta$}
		{
			\textbf{Return} $s_2$.
		}\Else{
			\uIf {$std_1 < std_2$}
			{
				\textbf{Return} $s_1$.
			}\uElseIf{$std_1 > std_2$}
			{
				\textbf{Return} $s_2$.
			}\Else
			{
				\textbf{Return} randon one from $\{s_1, s_2\}$.
			}
		}
	}

\end{algorithm}
We develop one slack version of the standard binary tournament selection, which are documented in Algorithm~\ref{alg_tournament_selection}, to select parent solutions for the crossover operations in the proposed EvoCNN method. Briefly, two sets of comparisons are employed. The comparisons between the mean values of individuals involves a threshold $\alpha$, and that comparisons between the parameter numbers involves another threshold $\beta$. If the parent solution cannot be selected with these comparisons, the individual with smaller standard derivation is chosen.

In practice, tremendous number of parameters exist in deep CNNs, which would easily cause the overfitting problem. Therefore, when the difference between the mean values of two individuals is smaller than the threshold $\alpha$, we further consider the number of connection weights. The slight change of the parameter numbers will not highly affect the performance of CNNs. Consequently, $\beta$ is also introduced.

By iteratively performing this selection, parent solutions are selected and stored into a mating pool. In the proposed EvoCNN method, the size of the mating pool is set to be the same of the population size.

\subsection{Offspring Generation}
\label{section_3_4}
The steps for generating offspring are given as follows:
\begin{enumerate}[step 1):]
	\item randomly select two parent solutions from the mating pool;\label{step_1}
	\item use crossover operator on the selected solutions to generate offspring;
	\item use mutation operator on the generated offspring;\label{step_3}
	\item store offspring, remove the parent solutions from the mating pool, and perform steps~\ref{step_1}-\ref{step_3} until the mating pool is empty.
\end{enumerate}

\begin{figure*}[!htp]
	\centering
	\subfloat[Unit Collection]{\includegraphics[width=1.6\columnwidth]{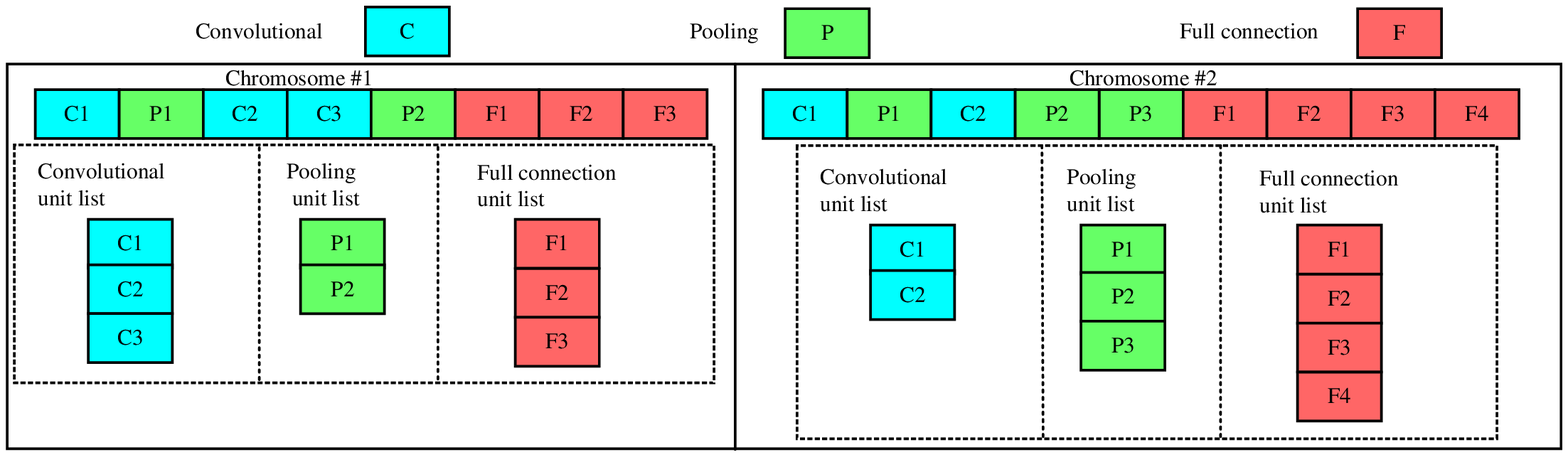}%
		\label{fig_first_crossover}}
	\hfil
	\subfloat[Unit Aligh and Crossover]{\includegraphics[width=1.6\columnwidth]{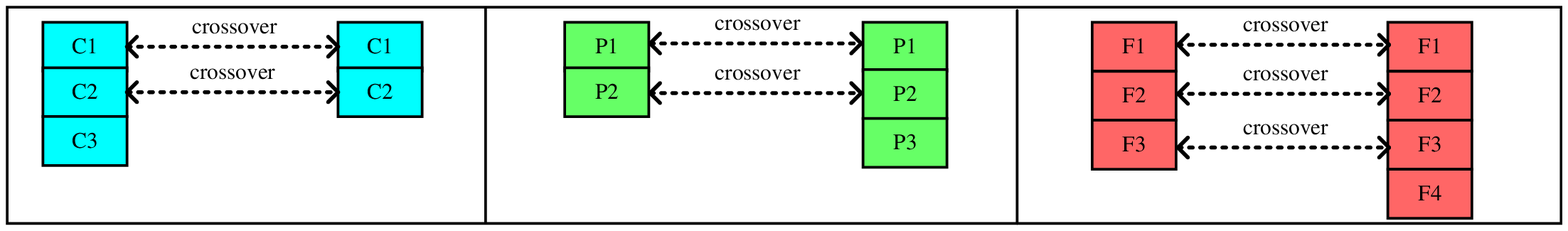}%
		\label{fig_second_crossover}}
	\hfil
	\subfloat[Unit Restore]{\includegraphics[width=1.6\columnwidth]{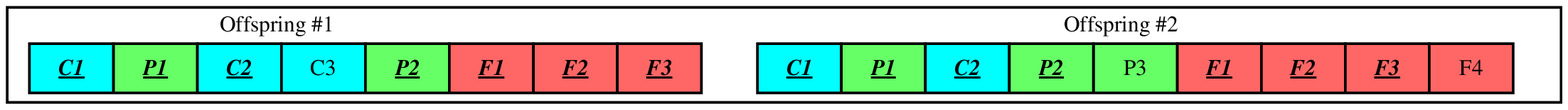}%
		\label{fig_third_crossover}}
	\caption{An example to illustrate the entire crossover process. In this example, the first chromosome is with length 8 including three convolutional layers, two pooling layers, and three full connection layers; the other one is with length 9 including two convolutional layers, three pooling layers, and four full connection layers. In the first step of crossover (i.e., the unit collection), the convolutiona layers, the pooling layers, and the full connection layers are collected from each chromosome and stacked with the same orders to them in each chromosome (see Fig.~\ref{fig_first_crossover}). In the second step, the unit lists with the same unit types are aligned at the top, i.e., the two convolutional layer lists are picked and aligned, and the same operations on the other two lists. When these unit lists finish the alignment, units at the same positions from the each two lists are paired and performed crossover operation, which can be shown in Fig.~\ref{fig_second_crossover}. At last, units from these unit lists are restored based on the positions where they are from (see Fig.~\ref{fig_second_crossover}). Note in Fig.~\ref{fig_second_crossover}, the units that have experienced crossover operations are highlighted with italics and underline fonts, while the units that do not perform the crossover operations remain the same.}
	\label{fig_crossover}
\end{figure*}

The proposed crossover operation can be seen in Fig.~\ref{fig_crossover}. 
To achieve crossover, we design a method called Unit Alignment (UA) for recombining two individuals with different chromosome lengths. During the crossover operation, three different units, i.e., the convolutional layer, the pooling layer, and the full connection layer, are firstly collected into three different lists based on their orders in the corresponding chromosome, which refers to the Unit Collection (UC) phase. Then, these three lists are aligned at the top, and units at the same positions are performed the crossover. This phase is named the UA and crossover phase. Finally, the Unit Restore (UR) phase is employed, i.e., when the crossover operation is completed, the units in these lists are restored to their original positions of the associated chromosomes. With these three consequential phases (i.e., the UC, the UA and crossover, and the UR), two chromosomes with different lengths could easily exchange their gene information for crossover. Because the crossover operation is performed on the unit lists where only the units with the same types are loaded, this proposed UA crossover operation is natural (because they have the same origins). For the remaining units, which do not perform crossover operations due to no paired ones, are kept at the same position.

Mutation operations may perform on each position of the units from one chromosome. For a selected mutation point, a unit could be added, deleted, or modified, which is determined by a probability of 1/3. In the case of unit addition, a convolutional layer, a pooling layer, or a full connection layer is added by taking a probability of 1/3. If the mutation is to modify an existing unit, the particular modification is dependent on the unit type, and all the encoded information in the unit would be changed (encoded information on each unit type can be seen in Table~\ref{tab_encoded_info}). Note that all the encoded formation is denoted by real numbers, therefore the Simulated Crossover (SBX)~\cite{deb1994simulated} and the polynomial mutation~\cite{deb2001multi} are employed in the proposed EvoCNN method due to their notable show in real number gene representations.

\subsection{Environmental Selection}
\label{section_3_5}
\begin{algorithm}
	\label{alg_environmental_selection}
	\caption{Environmental Selection}
	\KwIn{The elistsm fraction $\gamma$, and the current population $P_t\cup Q_t$.}
	\KwOut{The selected population $P_{t+1}$.}
	$a\leftarrow$ Calculate the number of elites based on $\gamma$ and the predefined population size $N$ from Algorithm~\ref{alg_population_init};\\
	$P_{t+1}\leftarrow$ Select $a$ individuals that have the best mean values from $P_t\cup Q_t$;\\
	$P_t\cup Q_t\leftarrow$ $P_t\cup Q_t - P_{t+1}$;\\
	\label{alg_es_removal_elites}
	\While{$|P_{t+1}|<N$}
	{\label{alg_es_diversity_begin}
		$s_1$, $s_2\leftarrow$ Randomly select two individuals from $P_t\cup Q_t$;\\
		$s\leftarrow$ Employe Algorithm~\ref{alg_tournament_selection} to select one individual from $s_1$ and $s_2$;\\
		$P_{t+1}\leftarrow$ $P_{t+1}\cup s$;\\
	}\label{alg_es_diversity_end}
	\textbf{Return} $P_{t+1}$.
\end{algorithm}

 The environmental selection is shown in Algorithm~\ref{alg_environmental_selection}. During the environmental selection, the elitism and the diversity are explicitly and elaborately addressed. To be specific, a fraction of individuals with promising mean values are chosen first, and then the remaining individuals are selected by the modified binary tournament selection demonstrated in Subsection~\ref{section_3_4}. By these two strategies, the elitism and the diversity are considered simultaneously, which are expected to collectively improve the performance of the proposed EvoCNN method.

 Note here that, the selected elites are removed before the tournament selection gets start to work (line~\ref{alg_es_removal_elites} of Algorithm~\ref{alg_environmental_selection}), while the individuals selected for the purpose of diversity are kept in the current population for the next round of tournament selection (lines~\ref{alg_es_diversity_begin}-\ref{alg_es_diversity_end} of Algorithm~\ref{alg_environmental_selection}) based on the convention in the community.

\subsection{Best Individual Selection and Decoding}
\label{section_3_6}
At the end of evolution, multiple individuals are with promising mean values but different architectures and connection weight initialization values. In this regard, there will be multiple choices to select the ``Best Individual''. For example, if we are only concerned with the best performance, we could neglect their architecture configurations and consider only the classification accuracy. Otherwise, if we emphasis the smaller number of parameters, corresponding decision could be made. Once the ``Best Individual'' is confirmed, the corresponding CNN is decoded based on the encoded architecture and connection weight initialization information, and then the decoded CNN will be deeply trained with a larger number of epochs by SGD for future usage.

\section{Experiment Design}
\label{section_4}
In order to quantify the performance of the proposed EvoCNN, a series of experiments are designed and performed on the chosen image classification benchmark datasets, which are further compared to state-of-the-art peer competitors. In the following, these benchmark datasets are briefly introduced at first. Then, the peer competitors are given. Finally, parameter settings of the proposed EvoCNN method participating into these experiments are documented.

\subsection{Benchmark Datasets}
In these experiments, nine widely used image classification benchmark datasets are used to examine the performance of the proposed EvoCNN method. They are the Fashion~\cite{xiao2017fashion}, the Rectangle~\cite{larochelle2007empirical}, the Rectangle Images (RI)~\cite{larochelle2007empirical}, the Convex Sets (CS)~\cite{larochelle2007empirical}, the MNIST Basic (MB)~\cite{larochelle2007empirical}, the MNIST with Background Images (MBI)~\cite{larochelle2007empirical}, Random Background (MRB)~\cite{larochelle2007empirical}, Rotated Digits (MRD)~\cite{larochelle2007empirical}, and with RD plus Background Images (MRDBI)~\cite{larochelle2007empirical} benchmarks.

 Based on the types of the classification objects, these benchmarks are classified into three different categories. The first category includes only Fashion, which is for recognizing 10 fashion objects (e.g., trousers, coats, and so on) in the $50,000$ training images and $10,000$ test images. The second one is composed of the MNIST~\cite{lecun1998gradient} variants including the MB, MBI, MRB, MRD, and the MRDBI benchmarks for classifying 10 hand-written digits (i.e., 0-9). Because the MNIST has been easily achieved $97\%$, these MNIST variants are arbitrarily added into different barriers (e.g., random backgrounds, rotations) from the MNIST to improve the complexity of classification algorithms. Furthermore, these variants have $12,000$ training images and $50,000$ test images, which further challenges the classification algorithms due to the mush less training data while more test data. The third category is for recognizing the shapes of objects (i.e., the rectangle or not for the Rectangle and RI benchmarks, and the convex or not for the Convex benchmark). Obviously, this category covers the Rectangle, the RI, and the CS benchmarks that contain $1,200$, $12,000$, and $8,000$ training images, respectively, and all of them include $50,000$ test images. Compared to the Rectangle benchmark, the RI is generated by adding randomly sampled backgrounds from the MNIST for increasing the difficulties of classification algorithms.

\begin{figure}[!htp]
 	\centering
 	\subfloat[Examples from the first category. From left to right, they are T-shirt, trouser, pullover, dress, coat, sandal, shirt, sneaker, bag, and ankle boot.]{\includegraphics[width=\columnwidth]{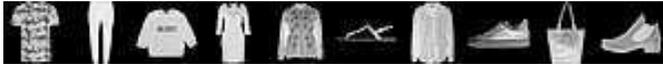}%
 		\label{fig_fashion_example}}
 	\hfil
 	\subfloat[Examples from the second category. From left to right, each two images are from one group, and each group is from MBi, MRB, MRD, MRDBI, and MB, respectively. These images refer to the hand-written digits 0, 4, 2, 6, 0, 5, 7, 5, 9, and 6, respectively. ]{\includegraphics[width=\columnwidth]{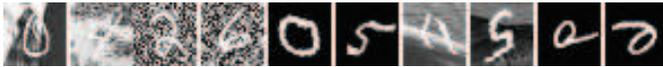}%
 		\label{fig_mnist_variant_example}}
 	\hfil
 	\subfloat[Examples from the thrid category, From left to right, the first three images are from the Rectangle benchmark, the following four ones are from the RI benchmark, and the remaingings are from the Convex benchmark. Specifically, these examples with the index 1, 2, 6, 7, and 11 are positive samples.]{\includegraphics[width=\columnwidth]{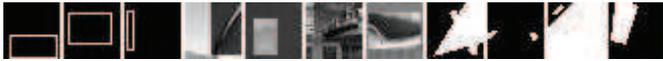}%
 		\label{fig_shape_example}}
 	\caption{Examples from the benchmarks chosen.}
 	\label{fig_benchmakrs}
    \vspace{-0.5mm}
 \end{figure}

 In addition, each image in these benchmarks is with the size $28\times 28$, and examples from these benchmarks are shown in Fig.~\ref{fig_benchmakrs} for reference. Furthermore, another reason for using these benchmark datasets is that different algorithms have reported their promising results, which is convenient for the comparisons on the performance of the proposed EvoCNN method and these host algorithms (details can be seen Subsection~\ref{subsection_peer_competitors}).

\vspace{-2.5mm}
\subsection{Peer Competitors}
\label{subsection_peer_competitors}
Ideally, the algorithms for neural network architecture design discussed in Section~\ref{section_1} should be collected here as the peer competitors. However, because MetaQNN~\cite{baker2017designing} and LEIC~\cite{real2017large} highly rely on the computational resources, it is impossible to reproduce the experimental results in an academic environment. Furthermore, the benchmark dataset investigated in these two algorithms employed different data preprocessing and augmentation techniques, which will highly enhance the final classification accuracy and is invisible to public. As for other architecture design approaches introduced in Section~\ref{section_1}, such as GS, RS, among others, they are not scalable to CNNs and not suitable to be directly compared as well.

In the experiments, state-of-the-art algorithms that have reported promising classification errors on the chosen benchmarks are collected as the peer competitors of the proposed EvoCNN method. To be specific, the peer competitors on the Fashion benchmark are collected from the dataset homepage\footnote{\url{https://github.com/zalandoresearch/fashion-mnist}}. They are 2C1P2F+Dropout, 2C1P, 3C2F, 3C1P2F+Dropout, GRU+SVM+Dropout, GoogleNet~\cite{szegedy2015going}, AlexNet~\cite{krizhevsky2012imagenet}, SqueezeNet-200~\cite{iandola2016squeezenet}, MLP 256-128-64, and VGG16~\cite{Simonyan14c}, which perform the experiments on the raw dataset without any preprocessing. The peer competitors on other benchmarks are CAE-2~\cite{rifai2011contractive}, TIRBM~\cite{sohn2012learning}, PGBM+DN-1~\cite{sohn2013learning}, ScatNet-2~\cite{bruna2013invariant}, RandNet-2~\cite{chan2015pcanet}, PCANet-2 (softmax)~\cite{chan2015pcanet}, LDANet-2~\cite{chan2015pcanet}, SVM+RBF~\cite{larochelle2007empirical}, SVM+Poly~\cite{larochelle2007empirical}, NNet~\cite{larochelle2007empirical}, SAA-3~\cite{larochelle2007empirical}, and DBN-3~\cite{larochelle2007empirical}, which are from the literature~\cite{chan2015pcanet} recently published and the provider of the benchmarks\footnote{\url{http://www.iro.umontreal.ca/~lisa/twiki/bin/view.cgi/Public/DeepVsShallowComparisonICML2007}}.

\subsection{Parameter Settings}
All the parameter settings are set based on the conventions in the communities of evolutionary algorithms~\cite{deb2002fast} and deep learning~\cite{hinton2012practical}, in addition to the maximal length of each basic layer. Specifically, the population size and the total generation number are set to be $100$. The distribution index of SBX and Polynomial mutation are both set to be 20, and their associated probabilities are specified as $0.9$ and $0.1$, respectively. The maximum lengths of the convolutional layers, the pooling layers, and the full connection layers are set to be the same as $5$ (i.e., the maximal depths of CNNs in these experiments are $15$). Moreover, the proportion of the elitism is set to be $20\%$ based on the Pareto principle. Note that the $20\%$ data are randomly selected from the training images as the fitness evaluation dataset. For the implementation of the proposed EvoCNN method, we constrain the same values of the width and height for filter, stride, and kernel. The width and height for stride in the convolutional layer are set to be $1$, those in the pooling layer are specified at the same value to its kernel, and the convolutional type is fixed to ``SAME''.

The proposed EvoCNN method is implemented by Tensorflow~\cite{abadi2016tensorflow}, and each copy of the code runs in a computer equipped with two GPU cards with the identical model number GTX$1080$. During the final training phase, each individual is imposed by the BatchNorm~\cite{ioffe2015batch} for speeding up and the weight decay with an unified number for preventing from the overfitting. Due to the heuristic nature of the proposed EvoCNN method, $30$ independent runs are performed on each benchmark dataset, and the mean results are estimated for the comparisons unless otherwise specified. Furthermore, the experiments take $4$ days on the Fashion benchmark for each run, $2$--$3$ days on other benchmarks. 

\section{Experimental Results and Analysis}
In this section, the experimental results and analysis of the proposed EvoCNN method against peer competitors are shown in Subsection~\ref{sec_over_results}. Then, the weight initialization method in the proposed EvoCNN method are specifically investigated in Subsection~\ref{sec_experiment_weight}.

\label{sec_experimental_results}
\subsection{Overall Results}
\label{sec_over_results}
Experimental results on the Fashion benchmark dataset are shown in Table~\ref{table_performance_mnist_fashion}, and those on the MB, MRD, MRB, MBI, MRDBI, Rectangle, RI, and Convex benchmark datasets are shown in Table~\ref{table_performance_mnist_variants}. In Tables~\ref{table_performance_mnist_fashion} and \ref{table_performance_mnist_variants}, the last two rows denote the \textit{best} and \textit{mean} classification errors received from the proposed EvoCNN method, respectively, and other rows show the \textit{best} classification errors reported by peer competitors\footnote{It is a convention in deep learning community that only the \textit{best} result is reported.}. In order to conveniently investigate the comparisons, the terms ``(+)'' and ``(-)'' are provided to denote whether the best result generated by the proposed EvoCNN method is better or worse that the best result obtained by the corresponding peer competitor. The term ``---'' means there is no available result reported from the provider or cannot be counted.  Most information of the number of parameters and number of training epoch from the peer competitors on the Fashion benchmark dataset is available, therefore, such information from EvoCNN is also shown in Table~\ref{table_performance_mnist_fashion} for multi-view comparisons. However, for the benchmarks in Table~\ref{table_performance_mnist_variants},  such information is not presented because they are not available from the peer competitors. Note that all the results from the peer competitors and the proposed EvoCNN method are without any data augmentation preprocessing on the benchmarks.

\begin{table}[!htp]
	\caption{The classification errors of the proposed EvoCNN method against the peer competitors on the Fashion benchmark dataset.}
	\label{table_performance_mnist_fashion}
	\renewcommand{\arraystretch}{1.2}
	\begin{center}
		\begin{tabular}{c|c|c|c}
			\hline
			\hline
			\textbf{classifier} & \textbf{error(\%)} & \textbf{\# parameters} & \textbf{\# epochs} \\
			\hline
			\hline
			2C1P2F+Drouout & 8.40(+) &3.27M & 300 \\
			\hline
			2C1P &7.50(+) &100K & 30\\
			\hline
			3C2F &9.30(+)& ---&--- \\
			\hline
			3C1P2F+Dropout & 7.40(+) &7.14M &150\\
			\hline
			GRU+SVM+Dropout & 10.30(+) &  ---&100\\
			\hline
			GoogleNet~\cite{szegedy2015going} & 6.30(+) & 101M& ---\\
			\hline
			AlexNet~\cite{krizhevsky2012imagenet} & 10.10(+) & 60M&--- \\
			\hline
			SqueezeNet-200~\cite{iandola2016squeezenet} & 10.00(+) & 500K & 200 \\
			\hline
			MLP 256-128-64 & 10.00(+) & 41K&  25\\
			\hline
			VGG16~\cite{Simonyan14c}& 6.50(+) & 26M & 200 \\
			\hline
			\cellcolor{Gray}EvoCNN (best)& \cellcolor{Gray}5.47 &\cellcolor{Gray}6.68M &  \cellcolor{Gray}100\\
			\hline
			\cellcolor{Gray}EvoCNN (mean)& \cellcolor{Gray}7.28 &\cellcolor{Gray}6.52M &  \cellcolor{Gray}100\\
			\hline
			\hline
		\end{tabular}
	\end{center}
\end{table}

It is clearly shown in Table~\ref{table_performance_mnist_fashion} that by comparing the best performance,  the proposed EvoCNN method outperforms \textit{all} the ten peer competitors. The two state-of-the-art algorithms, GoogleNet and VGG16, achieve respectively 6.5\% and 6.3\% classification error rates, where the difference is only 0.2\%.  The proposed EvoCNN method further decreases the error rate by 0.83\% to 5.47\%.  Furthermore, the mean performance of EvoCNN is even better than the best performance of eight competitors, and only a little worse than the best of GoogleNet and VGG16. However, EvoCNN has a much smaller number of connection weights --- EvoCNN uses 6.52 million weights while GoogleNet uses 101 million and VGG uses 26 million weights. EvoCNN also employs only half of the numbers of training epochs used in the VGG16.  The results show that the proposed EvoCNN method obtains much better performance in the architecture design and connection weight initialization of CNNs on the Fashion benchmark, which significantly reduces the classification error of the evolved CNN generated by the proposed EvoCNN method.

\begin{table*}[!htp]
	\caption{The classification errors of the proposed EvoCNN method against the peer competitors on the MB, MRD, MRB, MBI, MRDBI, Rectangle, RI, and Convex benchmark datasets}
	\label{table_performance_mnist_variants}
	\renewcommand{\arraystretch}{1.2}
	\begin{center}
		\begin{tabular}{c|c|c|c|c|c|c|c|c}
			\hline
			\hline
			\textbf{classifier} & \textbf{MB} & \textbf{MRD} & \textbf{MRB} & \textbf{MBI} &\textbf{MRDBI} &\textbf{Rectangle} &\textbf{RI} &\textbf{Convex} \\
			\hline
			\hline
			CAE-2~\cite{rifai2011contractive} & 2.48(+) & 9.66(+) &10.90(+) & 15.50(+) & 45.23(+) &1.21(+) &21.54(+) & ---\\
			\hline
			TIRBM~\cite{sohn2012learning} & --- & 4.20(-) & --- & --- & 35.50(+) & --- &  --- & --- \\
			\hline
			PGBM+DN-1~\cite{sohn2013learning} &  --- &  --- & 6.08(+) & 12.25(+)&36.76(+) & --- & --- & --- \\
			\hline
			ScatNet-2~\cite{bruna2013invariant} & 1.27(+) & 7.48(+) & 12.30(+)&18.40(+)& 50.48(+) & 0.01(=) & 8.02(+) & 6.50(+) \\
			\hline
			RandNet-2~\cite{chan2015pcanet} & 1.25(+) & 8.47(+)& 13.47(+) & 11.65(+) & 43.69(+) & 0.09(+) & 17.00(+) &5.45(+)\\
			\hline
			PCANet-2 (softmax)~\cite{chan2015pcanet}&1.40(+) & 8.52(+) & 6.85(+)& 11.55(+)& 35.86(+) & 0.49(+) & 13.39(+) & 4.19(-) \\
			\hline
			LDANet-2~\cite{chan2015pcanet} & 1.05(-) & 7.52(+) & 6.81(+) & 12.42(+) & 38.54(+) & 0.14(+) & 16.20(+) & 7.22(+) \\
			\hline
			SVM+RBF~\cite{larochelle2007empirical} & 3.03(+) & 11.11(+) & 14.58(+) & 22.61(+) & 55.18(+) & 2.15(+) & 24.04(+) &19.13(+) \\
			\hline
			SVM+Poly~\cite{larochelle2007empirical} & 3.69(+) & 15.42(+) & 16.62(+)& 24.01(+) & 56.41(+) &2.15(+) & 24.05(+) & 19.82(+) \\
			\hline
			NNet~\cite{larochelle2007empirical} & 4.69(+) & 18.11(+) & 20.04(+)& 27.41(+) & 62.16(+) & 7.16(+) & 33.20(+) & 32.25(+) \\
			\hline
			SAA-3~\cite{larochelle2007empirical} & 3.46(+) & 10.30(+) & 11.28(+)& 23.00(+) & 51.93(+) & 2.41(+) & 24.05(+) & 18.41(+) \\
			\hline
			DBN-3~\cite{larochelle2007empirical} & 3.11(+) & 10.30(+)& 6.73(+) & 16.31(+) & 47.39(+) & 2.61(+) & 22.50(+) & 18.63(+) \\
			\hline
			\cellcolor{Gray}EvoCNN (best) &\cellcolor{Gray}1.18 &\cellcolor{Gray}5.22 &\cellcolor{Gray}2.80 &\cellcolor{Gray} 4.53&\cellcolor{Gray} 35.03 &\cellcolor{Gray}0.01 &\cellcolor{Gray}5.03  &\cellcolor{Gray} 4.82\\
			\hline
			\cellcolor{Gray}EvoCNN (mean) &\cellcolor{Gray}1.28 &\cellcolor{Gray}5.46 &\cellcolor{Gray}3.59 &\cellcolor{Gray} 4.62&\cellcolor{Gray} 37.38 &\cellcolor{Gray}0.01 &\cellcolor{Gray}5.97  &\cellcolor{Gray} 5.39\\
			\hline
			\hline
		\end{tabular}
	\end{center}
\end{table*}

According to Table~\ref{table_performance_mnist_variants},  EvoCNN is the \textit{best} one among all the 13 different methods. Specifically, EvoCNN achieves the best performance on five of the eight datasets, and the second best on the other three datasets --- the MB, MRD, and Convex datasets, where LDANet-2, TIRBM, and PCANet-2(softmax) achieve the lowest classification error rate, respectively. Furthermore, comparing the \textit{mean} performance of EvoCNN with the \textit{best} of the other 12 methods,  EvoCNN is the best on four datasets (MRB, MBI,  Rectangle and RI), second best on two (MRD and Convex), third best and fourth best on the other two datasets (MRDBI and MB).  Particularly on MRB and MBI,  the lowest classification error rates of the others are 6.08\% and 11.5\%, and the mean error rates of EvoCNN are 3.59\% and 4.62\%, respectively.
In summary, the best classification error of the proposed EvoCNN method wins 80 out of the 84 comparisons against the best results from the 12 peer competitors, and the mean classification error of EvoCNN is better than the best error of the 12 methods on 75 out of the 84 comparisons.


\subsection{Performance Regarding Weight Initialization}
\label{sec_experiment_weight}
\begin{figure}[htp]
	\centering
	\includegraphics[width=\columnwidth]{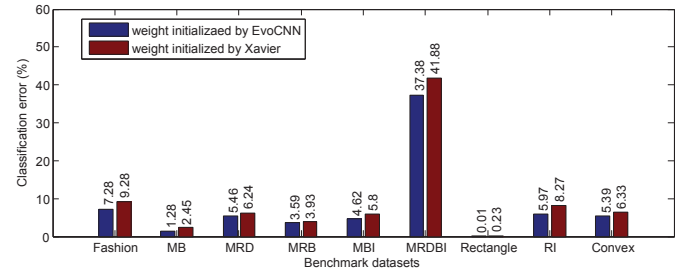}
	\caption{Performance comparison between the proposed EvoCNN method and the CNN using the evolved architecture and the Xavier weight initializer.}
	\label{fig_weight_comparision}
\end{figure}

Further experiments are performed to examine the effectiveness of the connection weight initialization method in the proposed EvoCNN approach. By comparing different weight initializers, we would also investigate whether the architecture or the connection weight initialization values will affect the classification performance. To achieve this, we initialize another group of CNNs with the same architectures as that of the evolved EvoCNN, but their weights are initialized with the widely used Xavier initializer~\cite{glorot2010understanding} (details in Subsection~\ref{subsection_weight_init_method}).

The comparisons are illustrated in Fig.~\ref{fig_weight_comparision}, where the x-axis denotes the benchmark datasets, and y-axis denotes the classification error rates. Fig.~\ref{fig_weight_comparision} clearly shows that the proposed connection weight initialization strategy in EvoCNN improves the classification performance on \textit{all} the benchmarks over the widely used Xavier initializer. To be specific, the proposed weight initialization method reaches $\approx$$1.5\%$ classification accuracy improvement on the Fashion, MB, MRD, MBI, RI, and Convex benchmarks, and $4.5\%$ on the MRDBI benchmark. By comparing with the results in Tables~\ref{table_performance_mnist_fashion} and \ref{table_performance_mnist_variants}, it also can be concluded that the architectures of CNNs contribute to the classification performance more than that of the connection weight initialization. The proposed EvoCNN method gained promising performance by automatically evolving both the initial connection weights and the architectures of CNNs.

\section{Further Discussions}
\label{section_5}
In this section, we will further discuss the encoding strategy of the architectures and the weights related parameters, and the fitness evaluation in the proposed EvoCNN method. Extra findings from the experimental results are discussed as well, which could provide insights on the applications of the proposed EvoCNN method.

In GAs, crossover operators play the role of exploitation search (i.e., the local search), and mutation operators act the exploration search (i.e., the global search). Because the local search and the global search should complement to each other, only well designing both of them could remarkably promote the performance. Generally, the crossover operators operate on the chromosomes with the same lengths, while the proposed EvoCNN method employs the variable-length ones. Furthermore, multiple different basic layers exist in CNNs, which improve the difficulties of designing crossover operation in this context (because crossover operations can not be easily performed between the genes from different origins). Therefore, a new crossover operation is introduced to the proposed EvoCNN method.  UC, UA, as well as UR are designed to complete the crossover operation, which enhances the communications between the encoded information, and expectedly leverages the performance in pursuing the promising architectures of CNNs.

The typically used approaches in optimizing the weights of CNNs are based on the gradient information. It is well known that the gradient-based optimizers are sensitive to the starting position of the parameters to be optimized. Without a better starting position, the gradient-based algorithms are easy to be trapped into local minima. Intuitively, finding a better starting position of the connection weights by GAs is intractable due to the huge numbers of parameters. As we have elaborated, a huge number of parameters can not be efficiently encoded into the chromosomes, nor effectively optimized. In the proposed EvoCNN method, an indirect encoding approach is employed, which only encodes the means and standard derivations of the weights in each layer. In fact, employing the standard derivation and the mean to initialize the starting position of connection weights is a compromised approach and can be ubiquitously seen from deep learning libraries~\cite{abadi2016tensorflow,jia2014caffe,chen2015mxnet}, which is the motivation of this design in the proposed EvoCNN method. With this strategy, only two real numbers are used to denote the hundreds of thousands of parameters, which could save much computational resource for encoding and optimization.

Existing techniques for searching for the architectures of CNNs typically take the final classification accuracy as the fitness of individuals. A final classification accuracy typically requires many more epochs of the training, which is a very time-consuming process. In order to complete the architecture design with this fitness evaluation approach, it is natural to employ a lot of computation resource to perform this task in parallel to speed up the design. However, computational resource is not necessarily available to all interested researchers. Furthermore, it also requires extra professional assistances, such as the task schedule and synchronization in the multi-thread context, which is beyond the expertise of most researchers/users. In fact, it is not necessary to check individual final classification accuracy, but a tendency that could predict their future quality would be sufficient. Therefore, we only employ a small number of epochs to train these individuals. With this kind of fitness measurement, the proposed EvoCNN method does not highly reply on the computational resource. In summary, with the well-designed yet simplicity encoding strategies and this fitness evaluation method, researchers without rich domain knowledge could also design the promising CNN models for addressing specific tasks in their (academic) environment, where is the computational resources are typically limited.

\begin{table}[!htp]
	\caption{The classification accuracy and the parameter numbers from the individuals on the Fashion benchmark dataset.}
	\label{table_performance_num_parameters}
	\renewcommand{\arraystretch}{1.2}
	\begin{center}
		\begin{tabular}{c|c|c|c|c|c}
			\hline
			\hline
			\textbf{classification accuracy} & 99$\%$ &  98$\%$ & 97$\%$ & 96$\%$ & 95$\%$ \\
			\hline
			\textbf{\# parameters} & 569,250 & 98,588 & 7,035 & 3,205 & 955\\
			\hline
			\hline
		\end{tabular}
	\end{center}
\end{table}

Moreover, interesting findings have also been discovered when the proposed EvoCNN method terminates, i.e., multiple individuals are with the similar performance but significant different numbers of connection weights (See Table~\ref{table_performance_num_parameters}), which are produced due to population-based nature of the proposed EvoCNN method. Recently, various applications have been developed, such as the auto-driving car and some interesting real-time mobile applications. Due to the limited processing capacity and battery in these devices, trained CNNs with similar performance but fewer parameters are much more preferred because they require less computational resource and the energy consumption. In addition, similar performance of the individuals are also found with different lengths of the basic layers, and there are multiple pieces of hardware that have been especially designed to speed the primitive operations in CNNs, such as the specialized hardware for convolutional operations. In this regard, the proposed EvoCNN method could give manufacturers more choices to make a decision based on their own preferences. If the traditional approaches are employed here, it is difficult to find out such models at a single run.

\section{Conclusions and Future Work}
\label{section_6}
The goal of the paper was to develop a new evolutionary approach to automatically evolving the architectures and weights of CNNs for image classification problems. This goal has been successfully achieved by proposing a new representation for weight initialization strategy, a new encoding scheme of variable-length chromosomes,  a new genetic operator, a slacked binary tournament selection, and an efficient fitness evaluation method. This approach was examined and compared with 22 peer competitors including the most state-of-the-are algorithms on nine benchmark datasets commonly used in deep learning. The experimental results show that the proposed EvoCNN method significantly outperforms all of these existing algorithms on almost all these datasets in terms of their best classification performance. Furthermore, the mean  classification error rate of the proposed algorithm is even better than the best of others in many cases.  In addition, the model optimized by EvoCNN is with a much smaller number of parameters yet promising best classification performance. Specifically, the model optimized by EvoCNN employs $100$ epochs to reach the lowest classification error rate of $5.47\%$ on the Fasion dataset, while the state-of-the-art VGG16 employs $200$ epochs but $6.50\%$ classification error rate. Findings from the experimental results also suggest that the proposed EvoCNN method could provide more options for the manufacturers that are interested in integrating CNNs into their products with limited computational and battery resources.

In this paper, we investigate the proposed EvoCNN method only on the commonly used middle-scale benchmarks. However, large-scale data are also widely exist in the current era of big data. Because evaluating only one epoch on these large-scale data would require significant computational resource and take a long period, the proposed fitness evaluation method is not suitable for them unless a huge amount computational resources are available. In the future, we will put effort on efficient fitness evaluation techniques. In addition, we will also investigate evolutionary algorithms for recurrent neural networks, which are powerful tools for addressing time-dependent data, such as the voice and video data.

\ifCLASSOPTIONcaptionsoff
  \newpage
\fi

\bibliographystyle{IEEEtran}
\bibliography{IEEEabrv,mybibfile}

\clearpage
\appendices

\end{document}